 \documentclass[pmlr,twocolumn,10pt]{jmlr} 

\usepackage{booktabs}
\usepackage{siunitx}

\usepackage{color}
\usepackage{comment}
\usepackage{multicol}
\usepackage{float}
\usepackage{multirow}

\usepackage{authblk} 

\usepackage{placeins}

\renewcommand{\copyright}{}
\newcommand{\tool}{\texttt{MoRE}}

 \title{MoRE: Multi-Modal Contrastive Pre-training with Transformers on X-Rays, ECGs, and Diagnostic Report}

\author[1]{Samrajya Thapa}
\author[2]{Koushik Howlader}
\author[3]{Subhankar Bhattacharjee}
\author[4]{Wei Le}

\affil[1]{Iowa State University, Email: svthapa@iastate.edu}
\affil[2]{Iowa State University, Email: howlader@iastate.edu}
\affil[3]{Iowa State University, Email: s7bhat@iastate.edu}
\affil[4]{Iowa State University, Email: weile@iastate.edu}
\date{} 

\begin{document}
\raggedbottom

\maketitle

\begin{abstract}
 In this paper,  we introduce a novel Multi-Modal Contrastive Pre-training Framework that synergistically combines X-rays, electrocardiograms (ECGs), and radiology/cardiology reports. Our approach leverages transformers to encode these diverse modalities into a unified representation space, aiming to enhance diagnostic accuracy and facilitate comprehensive patient assessments. We utilize LoRA-Peft to significantly reduce trainable parameters in the LLM and incorporate recent linear attention dropping strategy in the Vision Transformer(ViT) for smoother attention. Furthermore, we provide novel multimodal attention explanations and retrieval for our model. To the best of our knowledge, we are the first to propose an integrated model that combines X-ray, ECG, and Radiology/Cardiology Report with this approach. By utilizing contrastive loss, MoRE effectively aligns modality-specific features into a coherent embedding, which supports various downstream tasks such as zero-shot classification and multimodal retrieval. Employing our proposed methodology, we achieve state-of-the-art (SOTA) on the Mimic-IV, CheXpert, Edema Severity, and PtbXl downstream datasets, surpassing existing multimodal approaches. Our proposed framework shows significant improvements in capturing intricate inter-modal relationships and its robustness in medical diagnosis that establishes a framework for future research in multimodal learning in the healthcare sector.
 You can find the code for our experiments at: \href{https://github.com/svthapa/MoRE}{github/MoRE}.
\end{abstract}

\section{Introduction}
Self-supervised and multimodal pre-training \cite{wang2023large} are emerging research fields in Natural Language Processing (NLP), Computer Vision (CV), and the medical domain \cite{lin2023medical}. These methods use different types of data like images, text, audio, and signals to improve learning. In multimodal pre-training, we combine these data types from the same subject to enhance task performance. There are two main types of pre-training: supervised and self-supervised. Supervised pre-training uses labeled data to train models from start to finish, ensuring they learn specific responses. Self-supervised pre-training, on the other hand, relies on large amounts of unlabeled data, allowing the model to learn patterns and features on its own.
\noindent
In the medical field, particularly in radiology, various diagnostic modalities are employed to assess conditions affecting the heart, lungs, brain, and more. Common radiological tools include X-rays, MRI, and CT scans, while cardiological assessments might use ECG/EKG and echocardiograms. Typically, a clinician might start with a less expensive and more accessible modality like an X-ray, and progressively use more detailed and costly tools such as MRI and CT scans depending on the initial findings. Given the varying cost and availability of these technologies, it raises several pertinent research questions: Can we leverage more readily available and less expensive diagnostic tools effectively? How can we harness the rich, embedded information across these multiple modalities for enhanced diagnosis? Furthermore, understanding the generalization capabilities of integrating multiple diagnostic methods is crucial. This leads us to explore whether the decisions derived from such multimodal diagnostic strategies are reliable and how we might expand the use of available modalities to fully utilize all accessible information for diagnosis. \\
\noindent Building on this foundation, we introduce our simple and effective, \textit{\textbf{M}ulti-M\textbf{o}dal Contrastive Pre-Training Framework for X-\textbf{R}ays, \textbf{E}CGs, and Radiology/Cardiology Report} (\textbf{MoRE}). Our framework leverages tri-modal pre-training by combining image data (X-rays), signal data (ECGs), and textual data (diagnostic reports) from the same patients. Recent studies (\cite{kim2023tmt} \cite{yang2022diffsound} \cite{zhang2022opt}) have shown the effectiveness of increasing the number of modalities in research, and we aim to extend this by integrating the two most common and accessible diagnostic tools for chest-related conditions: X-rays and ECGs. These modalities complement each other, as the information missing in one is often present in the other. Inspired by the recent work ImageBind \cite{girdhar2023imagebind}, which connects different modalities through a common modality, our framework seeks to link the X-ray and ECG modalities via the textual modality of diagnostic reports. This makes it a unique tri-modal approach: X-ray, ECG, and Diagnosis Report, marking it the first initiative in the medical field to integrate these three modalities. \\
\noindent As we expand our framework to include the text modality, we encounter significant challenges in memory management due to the high memory consumption of large language models (LLMs). Operating three distinct models on a single GPU poses substantial difficulties. To mitigate these issues, we adopt the LoRa - PEFT strategy \cite{hu2023llmadapters}, which effectively reduces the trainable parameters of our LLM to just 0.6\% of its original size. This reduction not only facilitates more efficient memory usage but also improves model performance. Crucially, the results from our LoRa-pretrained LLM indicate a reduced susceptibility to catastrophic forgetting, further bolstering its utility in our multimodal approach.
\noindent{\bf Our contribution in this work can be noted as:}
\begin{itemize}
    \setlength{\itemsep}{2pt}
    \setlength{\parskip}{2pt} 
    \setlength{\topsep}{2pt} 
    \item Implement a unified tri-modal framework that is capable of diverse downstream task and perform at a high accuracy for all modality.
    \item Show the generalization capability of the pre-trained model through Zero-shot Classification 
    \item Provide multimodal explainability with Gradient-based Attention Visualization and multimodal retrieval.
\end{itemize}

\begin{figure*}[ht]
\centering
\includegraphics[width=\textwidth, height=7cm]{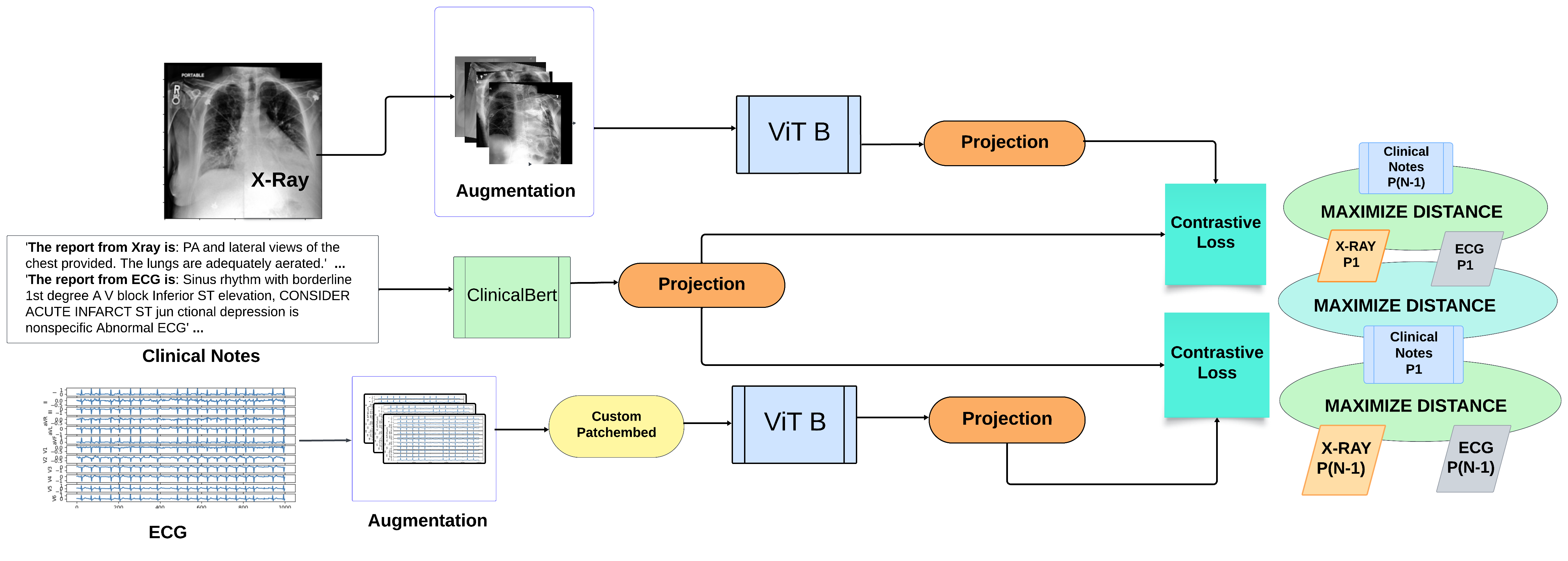}
\caption{MultiModal Pretraining Framework. We join the diagnostic report of both modalities as a single input and align the modalities with contrastive loss. We employ DropToken algorithm in our ViT encoders and custom patch embedding for ECG signal modality. The LLM is only fine-tuned with LoRA PEFT effectively training 0.6\% of its total parameters.}
\label{fig:overalldiagram}
\end{figure*}

\section{Related Work}
In this section we give background on our work, and introduce some of the recent works. 
\subsection{Vision Transformers}
The Vision Transformer (ViT) has recently set new benchmarks in several key areas, achieving state-of-the-art results in image classification on ImageNet, object detection on COCO, segmentation, and other tasks \cite{yin2023avit}. Traditionally, convolutional neural networks (CNNs) have been the go-to for such tasks because of their ability to handle increasingly complex patterns through larger kernels or receptive fields. The introduction of residual connections \cite{ebrahimi2023study} has allowed CNNs to grow significantly deeper, enhancing their ability to capture more complex information. 
Vision Transformers operate on a global scale using self-attention mechanisms. Although lacking the inductive biases of CNNs, given sufficient data, ViTs can learn intricate relationships within the data, offering a comprehensive understanding of the input.

\subsection{Self-Supervised Training}
Self-supervised pretraining \cite{liu2023selfsupervised}, has naturally gained popularity as the volume of available data increases while annotations or ground truths remain scarce. 
Recent advancements have shown that self-supervised learning not only reduces the dependency on labeled data but also improves the generalizability of models.

\subsection{Multi-Modal Pre-training}
Multimodal pretraining \cite{wang2023large}, establishes a unified framework that significantly enhances model generalization across various downstream tasks, such as classification, visual question answering (VQA), and segmentation. 
By capturing the fundamental features across different modalities during the pretraining phase, these models can, in some cases, outperform fully supervised models in downstream tasks. 
Moreover, the multimodal framework offers the flexibility to leverage all available modalities or select individual modalities for specific tasks, enhancing adaptability and application potential across a broader range of scenarios.

\subsubsection{Vison Language Pretraining in Medical Domain:}
Recent developments in Vision Language Pretraining (VLP) within the medical domain primarily utilizes contrastive learning, integrating image and text modalities. ConVIRT \cite{zhang2020contrastive} effectively pairs X-ray images with radiology reports, employing contrastive learning as its pretraining objective. Innovations have continued with GLoRIA \cite{huang2021gloria}, which not only uses X-ray and radiology reports but also introduces a novel architecture. This architecture captures both global and local features of each modality, utilizing `GloRIA Loss', a variation of contrastive loss, and traditional contrastive loss for pretraining. Furthermore, MedKLIP \cite{wu2023medklip} refines data processing by extracting a triplet of `Entity, Position, Exist' from radiology reports to eliminate irrelevant information and align textual entities with image patches spatially. In the context of ECG data, FrozenSSL \cite{li2023frozen} demonstrates the effectiveness of using a pre-trained, frozen ClinicalBert LLM with contrastive loss, avoiding additional fine-tuning for text modality integration.\\
Despite recent advancements, a significant gap remains in fully leveraging the broad range of available modalities for diagnosis. Recent work \cite{han2024fusemoemixtureofexpertstransformersfleximodal} integrates X-ray, ECG, and diagnostic reports but primarily addresses the challenges of missing modalities and modality collapse. In contrast, our work emphasizes the alignment of modalities while maintaining the capability to operate with single modalities. Similarly, \cite{wu2024multimodal} also tackles the issue of missing modalities but focuses on ICU data rather than downstream classification or zero-shot learning. Our paper proposes to integrate ECG data alongside its associated cardiology report—with existing X-ray data. This tri-modal approach—X-ray, ECG, and Radiology/Cardiology Report—aims to enhance the pretraining process, leveraging three modalities to improve both generalization and performance in downstream tasks and shows superior zero-shot performance.

\vspace*{-3mm}  
\section{Method}
In this section, we introduce our architecture, explain the different components, and the objective task.

\subsection{Architecture}
In our method, we utilize the Vision Transformer (ViT) as the backbone encoder for both the X-ray and ECG modalities. To enhance training stability and accelerate convergence, we initialize the ViT with pretrained ImageNet weights. The overall architecture is illustrated in Figure \ref{fig:overalldiagram}.


\subsection{Modality Encoder}


We implement a custom patch embedding to encode the ECG modality shown in Appendix \ref{custom} fig \ref{fig:ViTandEmbedding}. 

\noindent We adopt DropKey \cite{li2022dropkey} strategy in the ViT self-attention shown in Figure \ref{fig:ViTandEmbedding}. Instead of dropping the attentin weights, we randomly mask the key with a linear rate over the layers. This allows for a smoother attention plot and better robustness. 
\noindent For textual modality, we use an extended version \cite{lewis2020pretrained} of Clinical Bert \cite{alsentzer2019publicly}, which is a Roberta base pretrained on Pubmed and Mimic reports. LLMs can be costly to train, specially in multimodal setting, so we fine-tune the LLM with LoRA PEFT strategy \cite{hu2021lora}.
\noindent Additionally, we concatenate the diagnostic reports from both modalities of the same patient into a input, utilizing the existing separator token in the LLM's tokenizer. This enables the LLM to encode text from both modalities into a unified input. We provide additional details on our encoders in Appendix \ref{appendix:pretraining_data_detail}
\subsection{Contrastive Loss}
The pretraining framework's objective centers on using contrastive loss to effectively manage relationships between modalities. Following established contrastive learning practices, we project the outputs from the modality-specific encoders into a low-dimensional shared space. Specifically, we employ the InfoNCE loss \cite{oord2018representation}, which aims to minimize the distance between features of the textual modality and those of the X-ray modality from the same patient, and similarly between the textual modality and the ECG modality.
Conversely, it maximizes the distance between non-corresponding/negative, pairs. By concatenating texts from both modalities with a separator token during tokenization, the model learns to discern which portion of text corresponds to which modality. We assess this capability through retrieval tasks, confirming the model's effectiveness in distinguishing and correctly associating the textual inputs with their respective modalities.

\noindent
The loss function can be formulated as follows: 
{
\begin{equation}
\mathcal{L}_{\text{InfoNCE}} = -\left[\log \frac{\exp(\text{sim}(z_i, z_j) / \tau)}{\sum_{k=1}^{N} \exp(\text{sim}(z_i, z_k) / \tau)}\right]
\end{equation}
where:
\begin{itemize}
    \setlength{\itemsep}{0pt} 
    \setlength{\parskip}{0pt} 
    \setlength{\topsep}{0pt}  
    \item $\text{sim}(z_i, z_j)$ represents the similarity between the representations $z_i$ and $z_j$, measured using the cosine similarity.
    \item $\tau$ denotes a temperature scaling parameter that controls the separation of the distribution.
    \item $N$ is the batch size
\end{itemize}
}
\noindent
We ensure that the loss is symmetric, i.e. we maximize and minimize distance bi-directional for both modalities. We make the temperature parameter learnable and allow the model to find the best value to fit our data. The loss function for our framework can be defined as: 

{
\begin{equation}
\mathcal{L} = \frac{1}{2} \left( \mathcal{L}_{\text{Text-Xray}} + \mathcal{L}_{\text{Text-ECG}} \right)
\end{equation}
}

\noindent
where:
\noindent 
\begin{equation}
\small
\begin{aligned}
\mathcal{L}_{\text{Text-Xray}} = -\frac{1}{2} \Bigg(& \log \frac{\exp(\text{sim}(z_{\text{Text}}, z_{\text{Xray}}) / \tau)}{\sum \exp(\text{sim}(z_{\text{Text}}, z_k) / \tau)} \\
& + \log \frac{\exp(\text{sim}(z_{\text{Xray}}, z_{\text{Text}}) / \tau)}{\sum \exp(\text{sim}(z_{\text{Xray}}, z_k) / \tau)} \Bigg)
\end{aligned}
\end{equation}

\vspace{-3.00 mm}

\begin{equation}
\small
\begin{aligned}
\mathcal{L}_{\text{Text-Ecg}} = -\frac{1}{2} \Bigg(& \log \frac{\exp(\text{sim}(z_{\text{Text}}, z_{\text{Ecg}}) / \tau)}{\sum \exp(\text{sim}(z_{\text{Text}}, z_{k}) / \tau)} \\
& + \log \frac{\exp(\text{sim}(z_{\text{Ecg}}, z_{\text{Text}}) / \tau)}{\sum \exp(\text{sim}(z_{\text{Ecg}}, z_{k}) / \tau)} \Bigg)
\end{aligned}
\end{equation}

\BlankLine


{
\begin{itemize}
    \setlength{\itemsep}{0pt} 
    \setlength{\parskip}{0pt} 
    \setlength{\topsep}{0pt}  
    \item \(\mathcal{L}_{\text{Text-Xray}}\) represents the InfoNCE loss for text-image pairs in both directions.
    \item \(\mathcal{L}_{\text{Text-ECG}}\) represents the InfoNCE loss for text-signal pairs in both directions.
    \item \(z_{k}\) represents the representation of a negative sample for both modalities
\end{itemize}
}

\section{Evaluation}

In this section, we introduce our research questions, experimental setup, and the results. 

\subsection{Experimental design}
Our experimental framework is meticulously structured to address each research question (RQ), ensuring a comprehensive evaluation of our model:
\vspace{-2.00 mm}
\begin{itemize}
    \setlength{\itemsep}{2pt} 
    \setlength{\parskip}{2pt} 
    \setlength{\topsep}{2pt}  
    \item \textbf{RQ1: \noindent{\bf Is \tool\ able to effectively learn the representation for ECG and X-Ray?}} \\
    We assess the zero-shot classification capabilities of our model for both X-ray and ECG modalities. Additionally, we present t-SNE plots to visualize the features of our model compared to baseline models, providing a clear graphical representation of its performance.
    \item \textbf{RQ2: \noindent{\bf Can \tool\ be fine-tuned to perform downstream classification tasks accurately?}} \\
    Our pre-trained model is fine-tuned on downstream datasets. The results of this fine-tuning are presented in a tabular format, allowing for direct comparison of performance metrics across different datasets.
    \item \textbf{RQ3: \noindent{\bf Is Multimodal of X-ray and ECG applicable in medical domain ?}} \\
    We evaluate the retrieval performance of our model by comparing it with baseline models. This comparison helps illustrate the effectiveness of our model in retrieving relevant medical images and data based on query inputs.
    \item \textbf{RQ4: \noindent{\bf How does \tool\ compare to single model pre-training ?}} \\
    We compare our multimodal approach against single modality models to demonstrate the benefits of integrating multiple types of data. This comparison aims to highlight the enhanced performance and utility that multimodality brings to medical image analysis.
\end{itemize}

\subsection{Experimental setup}

\subsubsection{Datasets}
For pretraining, we use the matched subset of Mimic CXR and Mimic ECG dataset. We find about 45k matching patients, with combination of about 800k X-Ray and ECG data. We provide further details on the matching of the pretraining dataset in Appendix \ref{appendix:pretraining_data_detail}. For downstream tasks, we test our model on the CheXpert \cite{chexpert2019} dataset which contains 192k frontal X-Ray images, Edema Severity \cite{Liao2021Pulmonary} which has 7k data with severity level as classification, PtbXl \cite{PTBXL2022} dataset which has 21k ECG data. For X-Ray images we test our model on the labels: \textit{Atelectasis (AT), Cardiomegaly (CM), Edema (ED), and Pleural Effusion (PE)}. For ECG images we test on superclass labels: \textit{Normal (NORM), Hypertrophy (HYP), ST/ T Changes (STTC)}, and Myocardial Infarction (MI). We use these labels as our baselines have worked on the same. For Zero-shot we use CheXpert 5x200, and Mimic Zero-Shot subsets created from their original datasets. We provide additional details on our datasets in Appendix \ref{Appendix:Datasets}

\subsection{Implementation}
Our model employs the `ViT-Base-patch16-224' pre-trained on ImageNet, utilizing the Timm library for its adaptability. The base encoder features 12 transformer layers and 12 multi-head self-attention heads. We apply a custom patch embedding for ECG modalities and handle diagnostic reports with ClinicalBert’s tokenizer. The model uses a projection layer with two linear layers and leverages the InfoNCE loss, configured with a learnable temperature parameter. Training is optimized with the AdamW optimizer and Automatic Mixed Precision, using gradient accumulation to manage large batch processing. Detailed implementation specifics, including parameter settings and architecture modifications, are provided in the Appendix \ref{appendix:implementation}

\subsection{Baselines}
We compare our Multi-Modal Contrastive Pre-training Framework (MoRE) with several state-of-the-art multimodal pretraining frameworks in the medical domain, including GLoRIA \cite{huang2021gloria}, MedKlip \cite{wu2023medklip}, ECG AdvMasking \cite{bo2022pretraining}, and FrozenSSL \cite{li2023frozen}. For a fair comparison, we pretrain GLoRIA on the Mimic-IV Dataset, aligning with the dataset used for our own pretraining and that of MedKLIP. Similarly, we pretrain ECG AdvMask and FrozenSSL on the Mimic-IV ECG 800k dataset. Notably, since FrozenSSL does not specify a normalization process for ECG, we adopt the same normalization approach used in our work. ECG AdvMask being a single modality framework employs an autoencoder to generate masks for ECG during pretraining, we follow their outlined process.
\subsection{Metrics}
In the medical domain, relying solely on accuracy to evaluate model performance can be misleading. Accuracy typically involves a fixed threshold (often 0.5) for classification, which does not account for the uneven distribution of classes, the varying difficulty of diagnosing certain conditions, or the prevalence of different conditions within the dataset. Instead, we utilize the Area Under the Receiver Operating Characteristic (AUROC) as our primary metric. AUROC measures the model's ability to discriminate between classes at various threshold settings, providing a more comprehensive assessment of performance across different clinical scenarios \cite{ling2003auc}.
\noindent
For evaluating the retrieval experiment, we use Precision@K metric and evaluate the text retrieved as correct if it falls in the same class label as the original text.
\subsection{Results and Discussion}
\subsubsection{RQ1: Zero-shot Classification}
The zero-shot process is described in Appendix \ref{appendix:zero-shot} \\

\noindent{\footnotesize \textit{AC: Atelectasis, CM: Cardiomegaly, ED: Edema, PE: Pleural Effusion}} \\
\noindent{\footnotesize \textit{NORM: Normal, HYP: Hypertrophy, STTC: ST/ T Changes, MI: Myocardial Infarction}}

\begin{table}[H]
\centering
\renewcommand{\arraystretch}{1.5} 
\resizebox{\linewidth}{!}{ 
\begin{tabular}{l|cccccccc}
\hline
\multirow{2}{*}{Model} & \multicolumn{4}{c}{CheXpert} & \multicolumn{4}{c}{Mimic-IV} \\
\cline{2-9}
& AC & CM & ED & PE & AC & CM & ED & PE \\
\hline
GloRIA & 0.66 & 0.79 & \textbf{0.71} & 0.73 & 0.63 & 0.68 & \textbf{0.81} & 0.72 \\
\hline
MedKLIP & 0.62 & 0.62 & 0.62 & 0.58 & 0.53 & 0.70 & 0.67 & 0.61 \\
\hline
\tool\  & \textbf{0.75} & \textbf{0.85} & 0.65 & \textbf{0.83} & \textbf{0.79} & \textbf{0.72} & 0.68 & \textbf{0.80}  \\
\hline
\end{tabular}
}
\caption{Zero-Shot for X-Ray}
\label{tab:Zero-Shot-X-ray}
\end{table}
\vspace{-5.00 mm}
\begin{table}[H]
\centering
\renewcommand{\arraystretch}{1.5}
\resizebox{\linewidth}{!}{
\begin{tabular}{l|ccccc}
\hline
\multirow{2}{*}{Method} & \multicolumn{4}{c}{PtbXl} \\
\cline{2-6}
& NORM & STTC & MI & HYP & CD \\
\hline
FrozenSSL & 0.50 & 0.40 & 0.42 & \textbf{0.58} & 0.51  \\
\hline
\tool\ & \textbf{0.77} & \textbf{0.56} & \textbf{0.50} & 0.56 & \textbf{0.52}  \\
\hline
\end{tabular}
}
\caption{Zero-Shot for ECG (Superclass)}
\label{tab:Zero-Shot-ecg}
\end{table}

\vspace{-3.00mm}

\begin{figure*}[h]
  \centering
  \includegraphics[width=\textwidth]{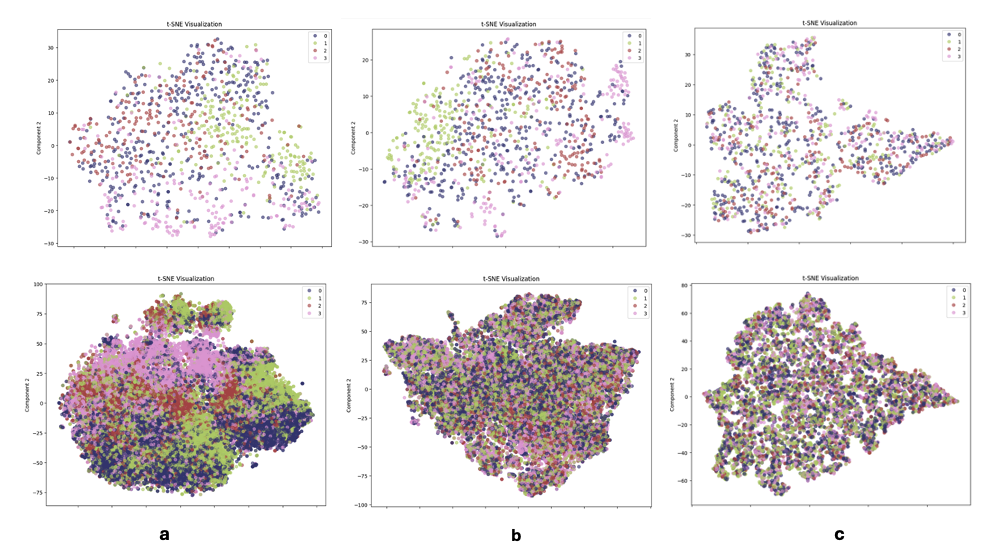}
  \caption{t-SNE plot of X-ray features of dataset Chexpert (top) and Mimic (bottom) from Models a: MoRE, b: GLoRIA, and c: MedKLIP.}
  \label{fig:tnseours}
\end{figure*}

\noindent
We also visualize the t-sne plot of the models features for X-Ray images shown in fig \ref{fig:tnseours}, which helps to illustrate the feature separations and clusters formed by different pathological labels. For the CheXpert dataset, we specifically utilize the subset designated as CheXpert 5x200. For the Mimic-IV dataset, we select a portion of the training data that contains images each uniquely labeled with one of the conditions, ensuring clarity in our visual analysis. This avoids any label conflicts in the plot. The conditions are labeled as follows: 0 for Atelectasis, 1 for Cardiomegaly, 2 for Edema, and 3 for Pleural Effusion. These t-SNE plots allow us to assess the distinctiveness of the model's feature representations for each pathology across the two datasets.

\BlankLine
\noindent{\bf Discussion}: 
\tool\ consistently outperforms GLoRIA and MedKlip in zero-shot classification for all labels except Edema. Notably, GLoRIA demonstrates superior performance specifically for Edema, as detailed in Table \ref{tab:Zero-Shot-X-ray}. The lower number of data points for Edema in the dataset may be a contributing factor to this performance gap. \tool\ outperforms FrozenSSL in all but one label by a small margin. 
\vspace{-2.00 mm}

\subsubsection{RQ2: Fine-tuning on downstream dataset}
We perform fine-tuning on our downstream datasets and report the results in tables \ref{X-ray-auc}, \ref{edema}, and \ref{ptbxl}  \\

\noindent
\begin{table}[H]
\centering
\resizebox{\columnwidth}{!}{ 
  \renewcommand{\arraystretch}{1.2}
  \begin{tabular}{l|cccccccc}
    \hline
    \multirow{2}{*}{Model} & \multicolumn{4}{c}{Mimic IV} & \multicolumn{4}{c}{CheXpert} \\
    \cline{2-9} 
    & AC & CM & ED & PE & AC & CM & ED & PE \\
    \hline
    GLoRIA & 0.73 & 0.73 & 0.81 & 0.84 & 0.76 & 0.77 & 0.87 & 0.90 \\
    \hline
    MedKLIP & 0.71 & 0.67 & 0.77 & 0.82 & 0.75 & 0.78 & \textbf{0.92} & \textbf{0.92} \\
    \hline
    \tool\ & \textbf{0.74} & \textbf{0.78} & \textbf{0.83} & \textbf{0.86} & \textbf{0.77} & \textbf{0.80} & \textbf{0.92} & 0.91 \\
    \hline
  \end{tabular}
}
\caption{X-ray Inference Results (AUC) in 100\% data}
\label{X-ray-auc}
\end{table}
\begin{table}[H]
\centering
\resizebox{0.8\columnwidth}{!}{ 
  \renewcommand{\arraystretch}{1.2}
  \begin{tabular}{l|cccc}
    \hline
    \multirow{2}{*}{Model} & \multicolumn{4}{c}{Edema Severity} \\
    \cline{2-5}
    & 0 & 1 & 2 & 3 \\
    \hline
    GLoRIA & 0.83 & 0.62 & 0.73 & 0.92 \\
    \hline
    MedKLIP & \textbf{0.85} & 0.66 & \textbf{0.76} & 0.88 \\
    \hline
    \tool\ & 0.82 & \textbf{0.78} & 0.75 & \textbf{0.92} \\
    \hline
  \end{tabular}
}

\caption{Edema Severity Results (AUC)}
\label{edema}
\end{table}

\begin{table}[H]
\centering
{\small
\begin{tabular}{l|ccccc}
\hline
Model & \multicolumn{5}{c}{PtbXL} \\
\cline{2-6} 
& Norm & STTC & HYP & MI & CD \\
\hline
AdvMask &0.89  &0.87  &0.85  &0.81  &0.84  \\
\hline
FrozenSSL &0.90  &0.86  &0.78  &0.77  &0.88  \\
\hline
\tool\ & \textbf{0.91} & \textbf{0.87} & \textbf{0.87} & \textbf{0.82} & \textbf{0.89} \\
\hline
\end{tabular}
}
\caption{PtbXL Superclass Results (AUC)}
\label{ptbxl}
\end{table}

\noindent \textbf{Discussion:} \tool\, outperforms both GLoRIA and MedKLIP on the Mimic-IV and CheXpert datasets across all labels but one. Notably, all models, including \tool\ , GLoRIA, and MedKlip, were trained exclusively on the Mimic-IV dataset. This context highlights the robustness and generalizability of \tool\ in handling diverse medical imaging data under similar training conditions. We also conduct downstream fine-tuning on Edema severity, to demonstrate that after fine-tuning, our framework achieves improved performance on the fine-grained classification task for Edema. Initially, Edema had shown lower performance in zero-shot classification, as detailed in Table \ref{tab:Zero-Shot-X-ray}. This fine-tuning highlights our framework's adaptability and effectiveness in enhancing performance on specific conditions that initially posed challenges.  Finally, \tool\ significantly outperformed both AdvMask and FrozenSSL as seen in Table \ref{ptbxl}, even though these models were trained on the entirety of the MImic-IV ECG dataset. In contrast, \tool\ was trained using only a matched subset of the Mimic-IV ECG dataset. This underscores \tool's efficiency and robustness in achieving high performance with less training data. 
\vspace{-3.00 mm}

\subsubsection{RQ3: Retrieval}
We conduct retrieval tasks to further validate the representations learned by our framework and demonstrate its application in medical learning. We utilize the CheXpert 5x200 dataset and a subset of the Mimic-IV dataset, where each X-ray is uniquely labeled with a diagnosis. This approach not only tests the effectiveness of the learned representations but also showcases how the framework can be applied in practical medical settings. 

\vspace{-1.00 mm}
\begin{table}[H]
\centering
\resizebox{\columnwidth}{!}{ 
  \begin{tabular}{l|cccccc}
    \hline
    \multirow{2}{*}{Model} & \multicolumn{3}{c}{Mimic-IV} & \multicolumn{3}{c}{CheXpert} \\
    \cline{2-7} 
    & Prec@5 & Prec@10 & Prec@100 & Prec@5 & Prec@10 & Prec@100 \\
    \hline
    GLoRIA & 27.0 & 26.2 & 26.7 & 27.7 & 27.7 & 26.7 \\
    \hline
    \tool\ & \textbf{52.9} & \textbf{51.0} & \textbf{44.2} & \textbf{55.8} & \textbf{55.6} & \textbf{50.0} \\
    \hline
  \end{tabular}
}
\caption{Precision@k Results for X-ray-to-Text}
\label{prec}
\end{table}
\raggedbottom

\noindent \textbf{Discussion:} We were unable to test for MedKLIP for this task because their approach does not involve training their LLM model; instead, they use it solely to encode the triplet extracted from the medical notes. We find our model performs better in retrieval for all K.
\BlankLine
\noindent{\bf Text-to-Xray Retrieval}
\noindent We demonstrate that, using a query text, our system can successfully retrieve X-rays associated with that query. The retrieval results display a variety of X-ray images that are relevant to the text. We believe this capability makes our tool highly valuable for educational purposes, helping users understand how a particular medical condition can appear in different patients, its various levels of severity, and the presence of comorbid conditions. This functionality could enhance learning and diagnostic training in medical education.\\
\noindent 
Example Query1: \textit{"Cardiomegaly is severe"}, Query2: \textit{"There is presence of Edema and Effusion"} 

\begin{figure}[!ht]
    \centering
    \begin{minipage}{0.45\textwidth}
        \centering
        \includegraphics[width=\textwidth]{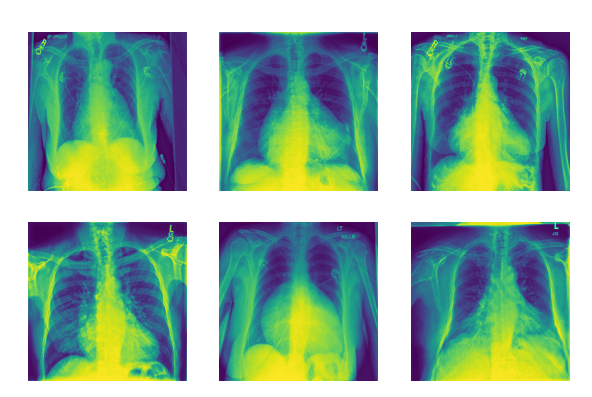}
        \caption{Retrieved X-ray of Query1: \textit{"Cardiomegaly is severe"}}
        \label{fig:imagesQuery1}
    \end{minipage}
    \hfill 
    \begin{minipage}{0.45\textwidth}
        \centering
        \includegraphics[width=\textwidth]{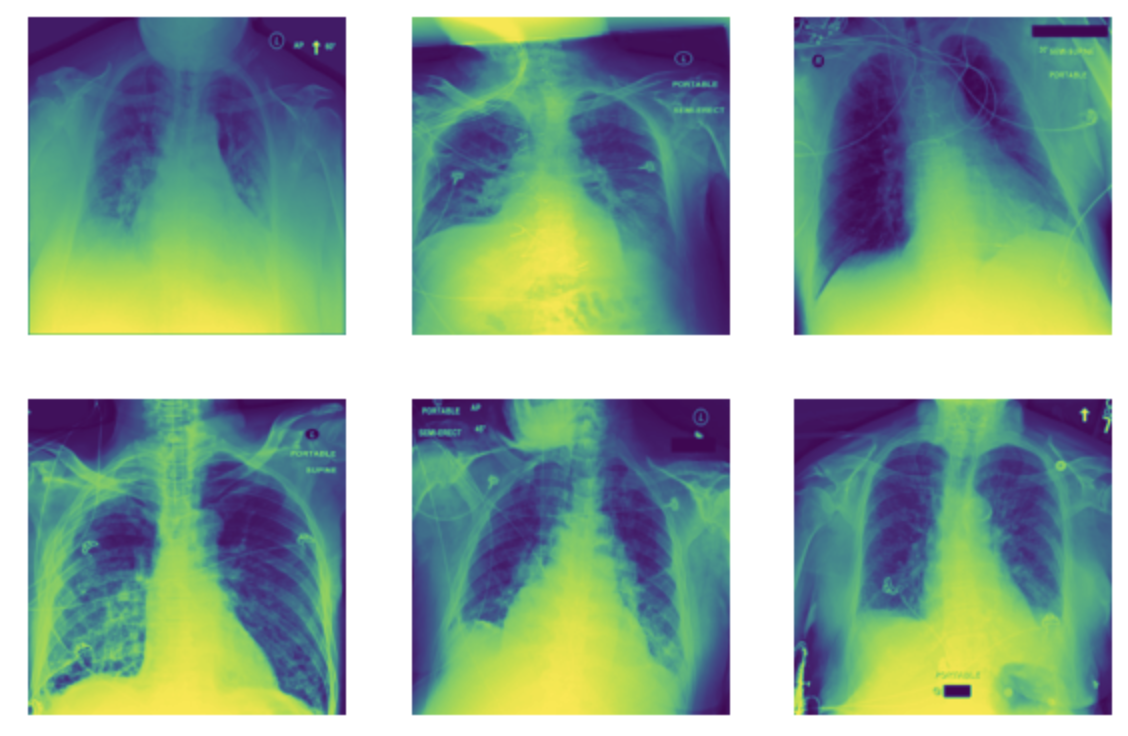}
        \caption{Retrieved X-ray of Query2: \textit{"There is presence of Edema and Effusion"}}
        \label{fig:imagesQuery2}
    \end{minipage}
\end{figure}
\raggedbottom

\begin{figure}[!ht]
  \centering
\includegraphics[width=0.55\textwidth]{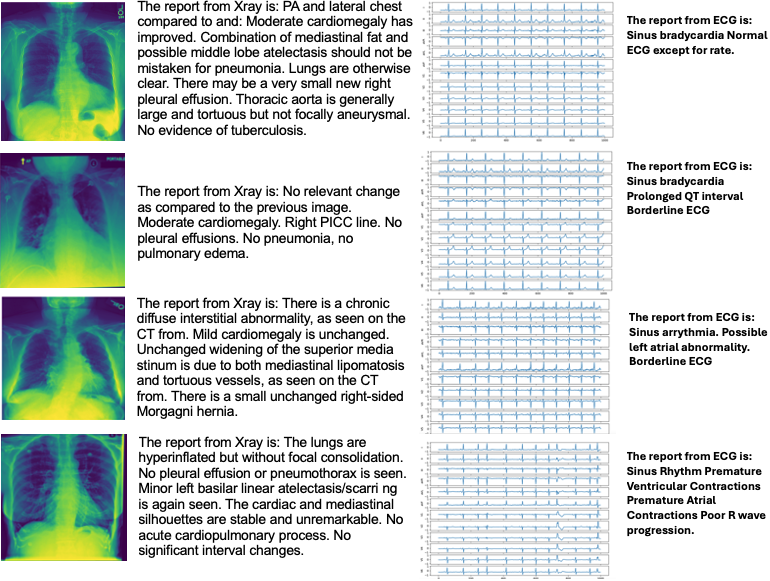}
  \caption{X-ray-ECG Retrieval with its original associated Text}
  \label{fig:imageX-rayecgtext}
\end{figure}

\vspace{0.1mm}
\noindent{\bf Discussion:} 
 In the retrieved images, the left X-ray images show a visibly enlarged heart, indicative of severe cardiac conditions. On the right, the X-ray images display fluid accumulation in both lung areas and at the base of the lungs, signaling the presence of edema and pleural effusion, as illustrated in Fig. \ref{fig:imagesQuery1}, \ref{fig:imagesQuery2}. This visual comparison highlights the tool's ability to effectively differentiate and display specific medical conditions critical for diagnostic purposes.
\BlankLine
\noindent{\bf MultiModal Retrieval}
We further demonstrate the capability of \tool\ to retrieve both X-ray and ECG data using a common text query in fig \ref{fig:imageX-rayecgtext}, showing that the text modality effectively binds the X-ray and ECG modalities together. This integration highlights \tool's ability to synthesize information across different types of medical data, offering a cohesive view that can be crucial for comprehensive diagnostic assessments.\\
\noindent{Example Query: \textit{"Cardiomegaly is present"}} 
\BlankLine


\noindent{\bf Discussion:} 
For the given query, the retrieved X-rays specifically mention Cardiomegaly in their original descriptions. Correspondingly, the retrieved ECGs show abnormalities such as "Arrhythmia, Bradycardia, Premature Ventricular Contractions(PVC)"—irregularities in heartbeat, slow heartbeat, and skipped or extra heartbeats, respectively, which are indicative of heart abnormalities. These results illustrate \tool's capability to align relevant diagnostic findings across modalities, enhancing the comprehensiveness of medical interpretations.
\BlankLine

\begin{figure*}[h]
  \centering
  \includegraphics[width=0.5\textwidth, height=4.5cm]{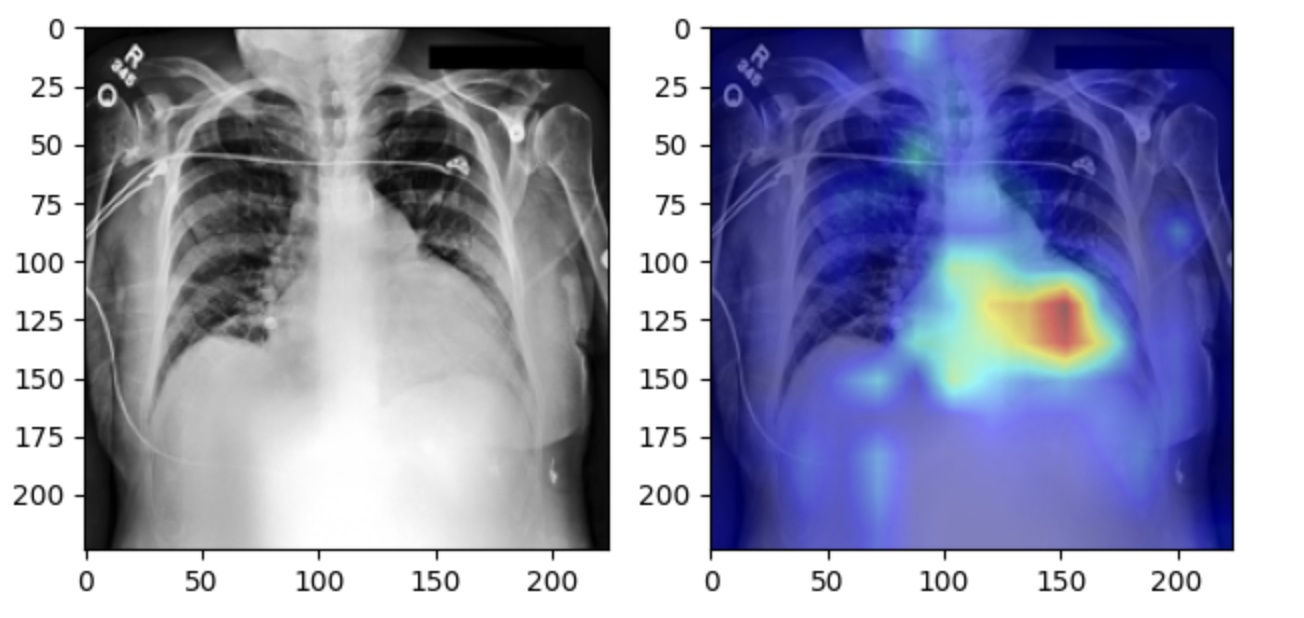}
  \hfill 
  \includegraphics[width=0.45\textwidth, height=6cm]{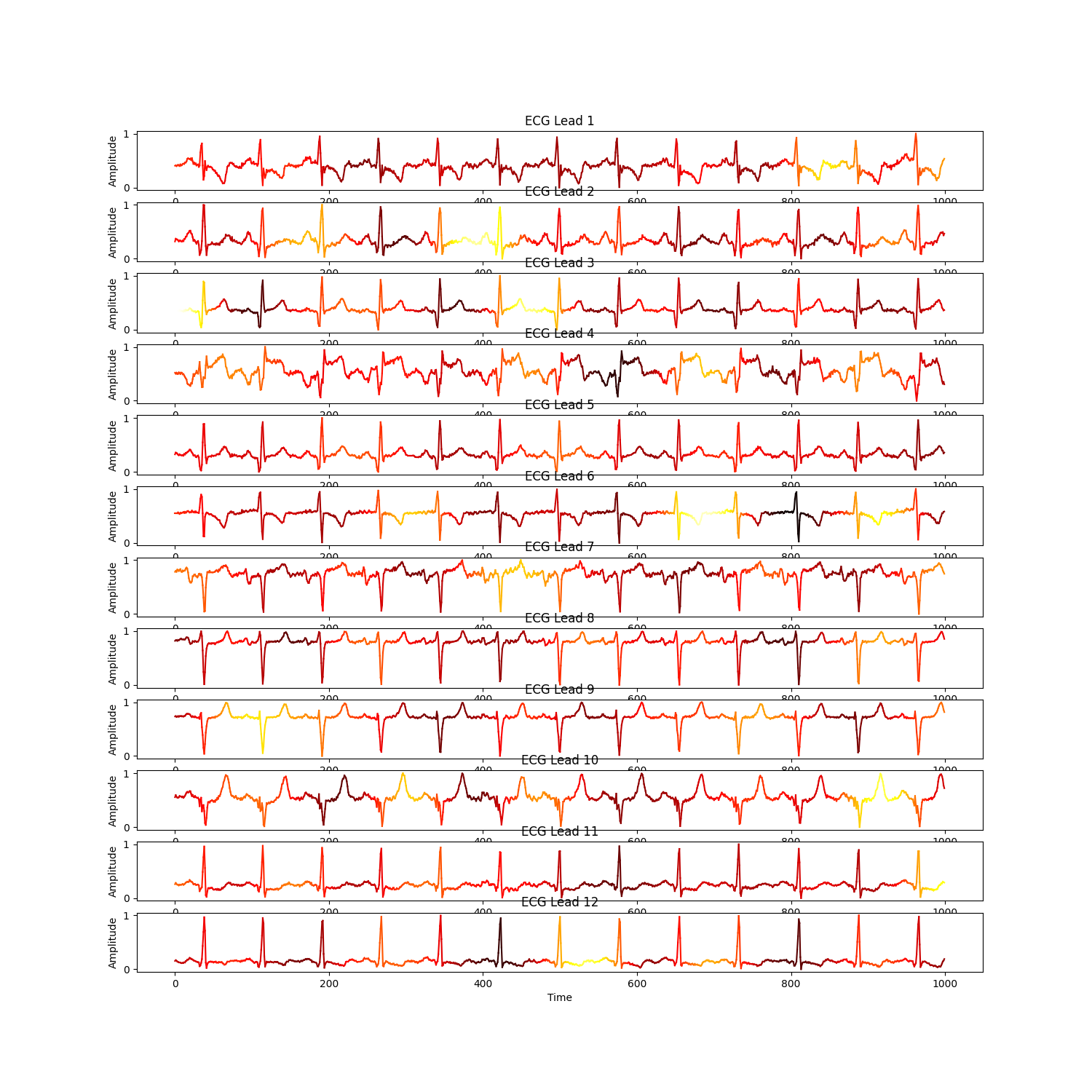}
  \caption{Left: Attn Plot on X-ray \quad \\ Right: Attn plot on ECG with Condition: Cardiomegaly}
  \label{X-ray and ecg vis}
  
  \includegraphics[width=0.45\textwidth]{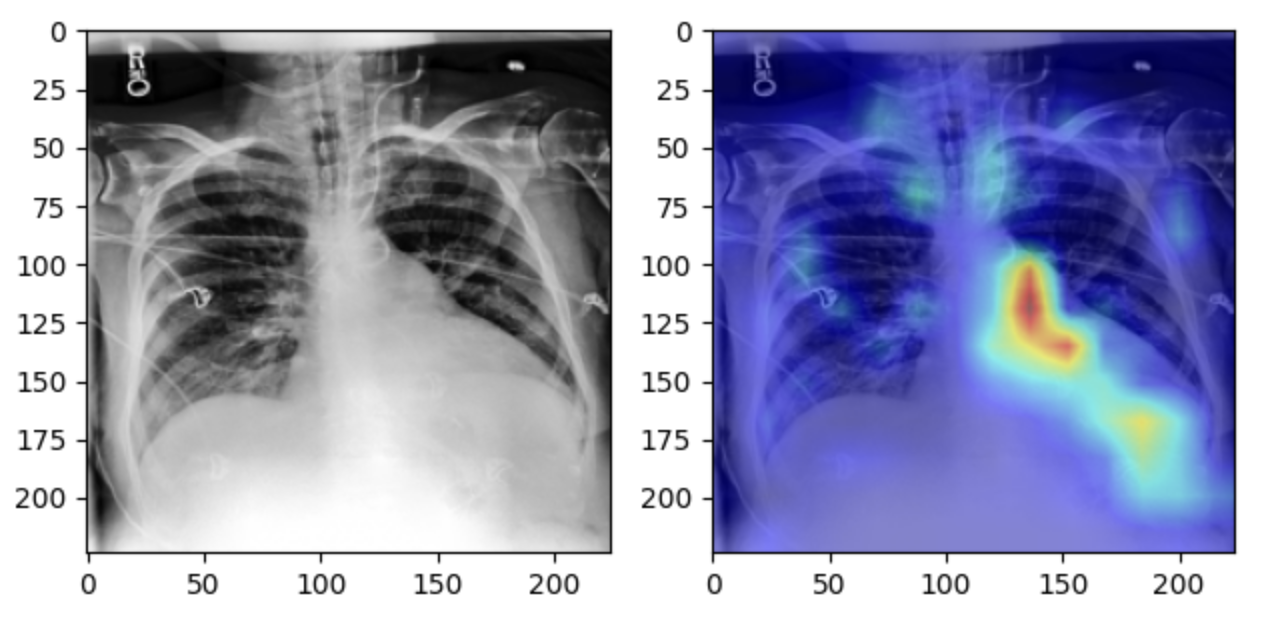}
  \hfill 
  \includegraphics[width=0.45\textwidth]{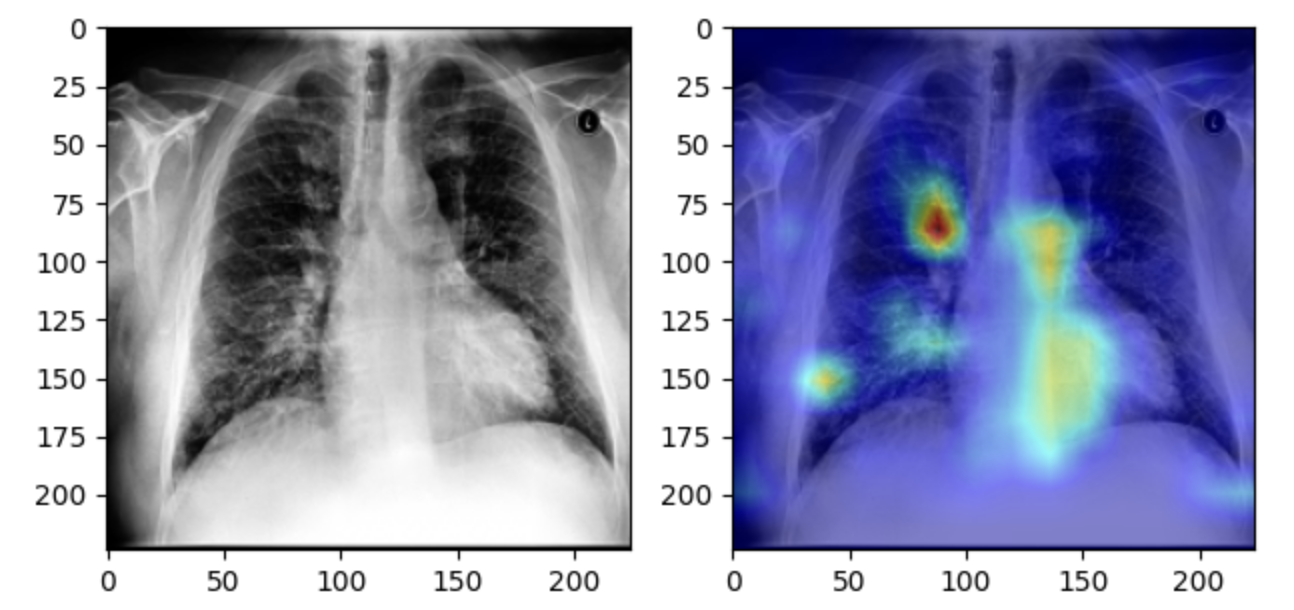}
  \caption{Left: X-ray with Cardiomegaly and Effusion \quad \\ Right: X-ray with Edema and Atelectasis}
  \label{X-ray vis}
\end{figure*}

\BlankLine
\noindent{\bf Gradient Based LRP Attention Visualization:}
We employ a modified version of \textit{TransLRP} \cite{byun2023vitreciproc}, which utilizes Layerwise Relevance Propagation (LRP) to aggregate backward gradient flow for deriving explanations. Our modification allows TransLRP to accept multimodal inputs, specifically X-ray and ECG, enabling us to visualize attention maps through the backward gradient flow. We present examples from the test set of the Mimic dataset unseen by the model, illustrating the attention focused on different diagnostic classes. This visualization helps clarify how the model prioritizes different aspects of the input data in making diagnostic decisions. 
We visualize the rollout attention of the transformer blocks based on the backward flow of the gradient from the classification head, which allows us to visualize the prominent attention for each class. In fig \ref{X-ray and ecg vis}, we are able to plot the attention on multimodal input of X-Ray and ECG when we use both modality for inference. For the given example, the condition is Cardiomegaly. We also show attention plot for X-Ray image with multi-class label in fig \ref{X-ray vis}.

\noindent{\bf Discussion:} In our analysis of the X-rays, we observe that the model directs high attention to areas of concern, specifically, the heart region for cardiomegaly and the base of the lungs for pleural effusion. For the ECGs, while the attention distribution is more complex to decipher, it is noticeably concentrated around the QRS complex, P Wave, and T Wave. These areas are crucial for identifying irregular heartbeats, which are prominent indicators of cardiac issues. This focused attention could serve as a valuable tool for clinical experts to validate the model's accuracy and relevance in real-world diagnostics, as demonstrated in Fig. \ref{X-ray and ecg vis} and \ref{X-ray vis}.





\subsubsection{RQ4 Results: Multimodal pretraining against Single modalities} 
We compare our model to single modality pre-training frameworks in Table \ref{multi}
We employ our base encoder, the ViT model, pre-trained on Mimic-IV data using contrastive learning with augmented data, similar to the approach used in SimCLR \cite{chen2020simple} with ImageNet initialization. We also benchmark against a fully supervised base ViT model to provide a comprehensive performance comparison. These models are evaluated using Mimic-IV data. For ECG, our baseline ECG AdvMask \cite{bo2022pretraining} is a single modality framework. Previous evaluations, as shown in Table \ref{ptbxl}, detail these comparisons, indicating how our multimodal approach stands against single modality training.

\begin{table}[!h]
\centering
{\small
\begin{tabular}{l|cccc}
\hline
\multirow{2}{*}{Model} & \multicolumn{4}{c}{Mimic IV} \\
\cline{2-5} 
& AC & CM & ED & PE\\
\hline
ViT-Base-IM(FS) &0.70 &0.76 &0.82 &0.83  \\
\hline
SimCLR &0.71 &0.74 &0.80 &0.81\\
\hline
\tool\ &\bf0.74 &\bf0.78 &\bf0.83 &\bf0.86\\
\hline
\end{tabular}
}
\caption{Single Model Pretraining Comparison with \tool\ . IM: Imagenet, FS: Fully-Supervised}

\label{multi}
\end{table} 
\BlankLine
\noindent \textbf{Discussion:}
We observe that \tool\ outperforms single-modality methods, such as its standalone ViT encoder when fully supervised and the ViT encoder pre-trained with the SimCLR framework. This highlights the superior fine-tuning capability of our pre-trained framework.

\vspace{-5.00mm}
\section{Ablation Study}
We study the use of incorporating all available modality for inference. Since Mimic IV dataset has a matched subset of X-Ray and ECG data, we perform inference with just X-Ray and X-Ray plus ECG. For the mulitmodal input, we ensure the study date of the X-Ray and ECG are no more than 3 days apart. 

\begin{table}[H]
\centering
{\small
\begin{tabular}{l|cccc}
\hline
\multirow{2}{*}{Modality} & \multicolumn{4}{c}{Mimic IV} \\
\cline{2-5} 
& AC & CM & ED & PE\\
\hline
X-Ray Only &0.74 &0.78 &0.83 &0.86  \\
\hline
X-Ray + ECG &0.73 &0.74 &0.76 &0.81\\
\hline
\end{tabular}
}
\caption{AUROC scores of \tool\ on Mimic IV}
\end{table}

\begin{table}[H]
\centering
{\small
\begin{tabular}{l|cccc}
\hline
\multirow{2}{*}{Modality} & \multicolumn{4}{c}{Mimic IV} \\
\cline{2-5} 
& AC & CM & ED & PE\\
\hline
X-Ray Only &0.46 &0.51 &0.53 &0.71  \\
\hline
X-Ray + ECG &0.50 &0.54 &0.65 &0.71\\
\hline
\end{tabular}
}
\caption{AUPRC scores of \tool\ on Mimic IV}
\end{table}

\noindent{\bf Discussion} Our results indicate that the AUROC scores are higher when utilizing only the X-ray modality. However, the AURPC scores improve when both the X-ray and ECG modalities are combined, suggesting that the model may be more robust to false negatives when incorporating additional information from the ECG data. Further investigation is needed to explore whether leveraging all available modalities can enhance diagnostic accuracy and improve overall model performance.

\vspace{4mm}
\noindent{\bf Limitations} \\
Our research effectively integrates multimodal data sources like X-rays, ECGs, and Diagnostic report but faces several limitations that affect its broader usability and effectiveness. Although the model shows strong performance on specific datasets such as Mimic-IV and CheXpert which, its generalizability to other datasets needs to be further tested. While LoRA PEFT strategy significantly reduces trainable parameters, we still cannot use larger available language models in a multimodal setting. Consequently, we were unable to utilize large language models (LLMs) like MEDITRON-7B and MEDITRON-70B \cite{chen2023meditron} due to GPU memory constraints.
\vspace{-4.00 mm}
\section{Conclusion}
This study successfully demonstrates the potential of our Multi-Modal Contrastive Pre-training Framework (MoRE) in enhancing diagnostic accuracy by integrating X-rays, ECGs, and clinical notes. Utilizing state-of-the-art transformer architectures and contrastive loss techniques, our model has shown superior performance on various benchmarks, establishing a new standard for multimodal learning in healthcare. Looking ahead, our future work will focus on overcoming the current limitations by expanding our evaluations to additional baseline datasets. We also plan to explore the use of large language models (LLMs) that could further extend the applicability and accessibility of our approach in different healthcare contexts.
\newpage
\bibliography{references}

\appendix

\section{DropKey and Custom Patch Embedding}
We use a custom patch embedding with 2 convolution layers each with ReLU activation and Batch Normalization to encode the ECG data before feeding it to the transformer model. We adopt DropKey in our transformer model which uses a linear drop rate to mask the Key of the transformer layers. The lower layers have a higher mask rate which linearly decreases to have the least masking rate at the last level. This is done to preserve the high-level information in the final layers.
\label{custom}
\begin{figure*}[!ht]
  \centering
  \includegraphics[width=0.40\textwidth, height=6.5cm]{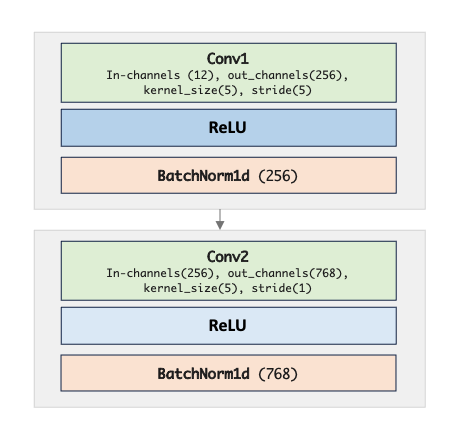}
  \includegraphics[width=0.40\textwidth, height=6.5cm]{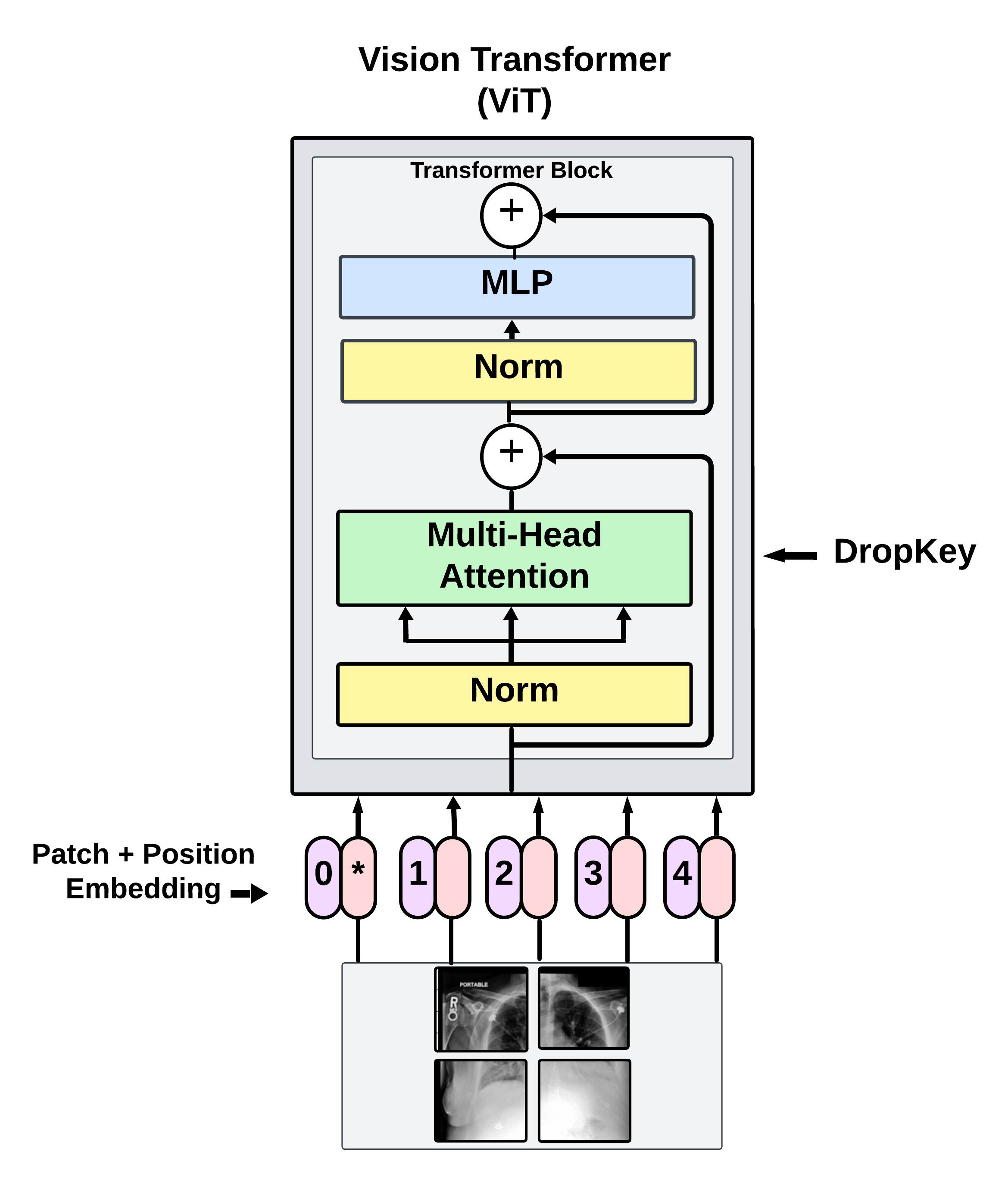}
  \caption{Left: Custom Patch Embedding for ECG \hspace{1cm}Right: ViT with DropKey} 
  \label{fig:ViTandEmbedding}
\end{figure*}
\vspace{-4.0mm}
\section{Pretraining Data Detail}
\label{appendix:pretraining_data_detail}
For Pre-training, we use two Mimic-IV dataset \cite{johnson2023mimiciv}, Mimic-CXR v2 \cite{johnson2019mimiccxr} and Mimic-IV ECG \cite{gow2023mimicivecg}. We use the permutation of the matched subset of Xray and ECG data from Mimic-IV and Mimic-CXR dataset. The total data we use for pre-training is about 800k where each data point is an Xray, ECG data, and Clinical Note from the same patient taken within 60 days. We ensure we only pretrain on \textit{`train'} fold of the dataset and leave out the validation and test set for downstream evaluation.\\
\noindent{\bf X-Ray Dataset:} \\
The Mimic-CXR-JPG dataset v2.0.0 \cite{johnson2024mimiccxrjpg} was released in Physionet. The dataset includes JPG images of chest Xray along with associated label and diagnostic text. The dataset has 227,827 Chest Xray images. The major reason for choosing this dataset is having matching ECG data of the patients in Mimic-IV ECG dataset \cite{gow2023mimicivecg} for the purposes of multimodal pretraining. \\
\noindent{\bf ECG Dataset:} \\
Mimic-IV ECG Matched Subset dataset \cite{gow2023mimicivecg} released in Physionet including \textit{800,000} ECG from \textit{160,000} patients. The dataset is derived from the larger Mimic-IV Clinical dataset which is also the parent of the Mimic-CXR dataset. The dataset contains no labels but includes clinical text. \\
\noindent{\bf Matched Subset:} \\
An individual Xray and ECG data item can be identified by some id's. To get the matched subset of data we identify an Xray and ECG through the patient's Sujbect ID, and Study ID. The subject id is a identifier of a patient, we find about little above 45k patients that have both Xray and ECG data present in the matched subset. There are multiple studies of the same patient taken in multiple dates, so we find a total of little more than 300k data points (Xray + ECG) creating this matched dataset of Xray, and ECG. We also add the clinical texts of the Xray and ECG that are available. For Xray data that did not have clinical notes associated with it, we create its note through its diagnosis label, e.g: \textit{ \"Finding of \{diagnosis\}, Uncertain Finding of \{diagnosis\}"}We follow this format because the clinical notes are in format of "\textit{Impression:}" and "Finding:". We add "\textit{uncertain finding}" for the Xray data points that have "\textit{-1}" in their label diagnosis, which is reported as uncertain finding in the dataset. \\
\noindent{\bf Clinical Notes:} \\
For the clinical notes of Xray, we use text under the headings in format of "Impression:", and "Findings:". We filter out the text from these headings and remove any special characters and redundant spaces from the text. For ECG, there are multiple reports for a ECG data, we merge the first 7 reports as we find them to be available for most ECGs and contain the most information. We then remove any special characters and redundant spaces from the text. \\
\noindent
During tokenization of the clinical notes, we tokenize the Xray and ECG note together with a separator token existing in the tokenizer of the LLM. 
\noindent{During matching Xray and ECG data of the same patient we carefully follow the following steps: }
\begin{enumerate}
    \setlength{\itemsep}{2pt} 
    \setlength{\parskip}{2pt} 
    \setlength{\topsep}{2pt} 
    \item Find all permutations of Xray and ECG studies of same patient
    \item Filter the data points if the de-identified dates of Xray and ECG are within 60days of each other. 
    \item Create Note of Xray data that do not have clinical note with its associated diagnosis. i.e. "\textit{Finding of \{diagnosis\}}"
\end{enumerate}

\section{Datasets}
\label{Appendix:Datasets}
\textbf{Pre-training Data}
For pretraining, we use matched subset from Mimic-CXR v2 \cite{johnson2019mimiccxr} and Mimic-IV ECG \cite{gow2023mimicivecg}. Mimic-CXR v2 has about 224k frontal chest X-ray images and Radiology Report, and Mimic-IV ECG has about 800k ECG signals with short Cardiology Report. There are about 45k matched subset of X-ray and ECG from the same patients. 
\BlankLine
\noindent\textbf{Datasets for Downstream Tasks} \\
\noindent{\bf Mimic-CXR} \cite{johnson2019mimiccxr} We use the Mimic-CXR v2 dataset for evaluating the pretrained representations on the test fold that is suggested by the dataset. The dataset has 13 labels but we choose to work on \textit{CM: Cardiomegaly, AT: Atelectasis, ED: Edema, and PF: Pleural Effusion}. We choose these labels as our baselines have also worked on them. Each X-ray data can have multiple diagnosis, making it a multi-label classification task.
\BlankLine\noindent
\noindent{\bf CheXpert} \cite{chexpert2019} We use the CheXpert dataset that has 192k X-ray data from 65k patients. We only utilize the frontal X-ray views and randomly sample \textit{5\%} of the training data for validation, and use the validation set of \textit{202} X-ray images for testing. This dataset does not come from the Mimic-IV parent corpus so this will be an outside dataset for our evaluation. We measure the AUROC which is presented in Table \ref{X-ray-auc}
\BlankLine\noindent
\noindent{\bf CheXpert 5x200} CheXpert 5x200 is taken from the CheXpert dataset such that each data point has only one unique diagnosis label. We utilize this dataset for Zero-Shot classification, precision@k, and retrieval tasks to prove the representation learned from our framework. 
\BlankLine\noindent
\noindent{\bf Mimic-Zero-Shot} Mimic-Zero-Shot is taken from the Mimic dataset such that each data point has only one unique diagnosis label. We utilize this dataset for Zero-Shot classification, precision@k, and retrieval tasks.
\BlankLine\noindent
\noindent{\bf Edema Severity} \cite{Liao2021Pulmonary} This dataset comes from the Mimic-CXR dataset. It contains about 7k data of which 6.6k are of training data, 520 are validation, and 140 additional for test. This split is as given in the dataset. The validation and test set are validated by multiple domain experts. The severity goes from 0, 1,2, and 3. \textit{0, none; 1, vascular congestion; 2, interstitial edema; and 3, alveolar edema}
\BlankLine\noindent
\noindent{\bf PtbXl ECG} \cite{PTBXL2022} This dataset that has 21k ECG data from 18k patients. Each ECG is associated with a diagnositc superclass label, namely: \textit{NORM : Normal ECG, HYP: Hypertrophy, STTC: ST/T changes, MI: Myocardial Infarction}
\vspace{-5.0mm}
\section{Data Pre-processing and Transformations}
\label{appendix:data-preprocessing}
Pre-processing is vital to deep learning model training as it can significantly impact the training times, convergence, and outcome. For medical data, it is key to ensure no important features are lost and conversely, highlight the important features. \\

\noindent{\bf Xray:}\\
For pre-processing we use \textbf{Adaptive Histogram Equalization} \cite{liu2023adaptive} to bring out the contrast and separation in the features. We find this step particularly important since the Xray images are greyscale and sometimes the quality of Xray varies introducing noise in the image and blending the features. After histogram equalization, the pixels are defined sharper and bring out features visibly. We then find the mean and standard deviation of the training dataset and use it for \textbf{Normalization} after transformations to improve training stability and performance. To bring some variability in the data, we follow work of \cite{van2024exploring} and use \textbf{RandomResizedScaling} with less stronger scaling of \textit{0.6-0.9} with \textit{0.8} probability, \textbf{RandomColorJitter} with brightness and contrast values of upto \textit{0.4} with \textit{0.8} probability, and \textbf{RandomGaussianBlur} with kernel size of \textit{7, 23} and with \textit{0.5} probability. \\

\noindent{\bf ECG:} \\
We follow a few key pre-processing steps for ECG. \textbf{1.Re-Sampling}: We resample the ECG data from 500Hz to 100Hz, changing its channel dimension from 5000 to 1000, making it a matrix of \textit{12,1000}. \textbf{2.Remove Nan}: We find there are NA values present in the data so we swap any NA values with 0. \textbf{3.Baseline Wander}: Baseline Wander is one of the pre-processing steps that absolutely cannot be missed. \cite{xu2017ecg} Baseline Wander is a low-frequency noise that is caused by the movement of the ECG leads. If not addressed can cause the model to deviate from the understanding of the data. Finally, we do \textbf{4.Per Lead Normalization} with MinMax Scaling to bring the range of each lead to -1 and 1. 
For Transformation, we follow work of \cite{raghu2022data} and utilize two augmentations for variability, \textbf{Time Warping}: warp \textit{4} segments of ECG by factor of \textit{0.25}, and \textbf{Random Permutation}: permute 4 segments of ECG in random order. 
\vspace{-3.00 mm}

\section{Implementation Detail}
\label{appendix:implementation}
Our implementation utilizes the `ViT-Base-patch16-224' model, pre-trained on ImageNet from the Timm library \cite{rw2019timm}, chosen for its ease of customization compared to the Hugging Face implementation. Our ViT base encoder consists of 12 transformer layers and 12 multi-head self-attention heads. We do not include any bias for query, key, value (qkv) during the linear projection of patches due to batch normalization after each layer.
\noindent
For the ECG modality, we employ a custom patch embedding strategy using two 1D-convolution layers, each followed by batch normalization and ReLU activation. The remainder of the ViT architecture is unchanged.
\noindent
The diagnostic reports are processed using ClinicalBert’s tokenizer with the parameter `add special tokens' $=$ True to insert separator tokens between texts from the two modalities. We set `max length' $=$ 512 and `truncation' $=$ True.
\noindent
Our projection layer comprises two linear layers with a hidden dimension of 768 and an output dimension of 128. The first linear layer is bias-free, followed by batch normalization and ReLU activation, while the second layer includes a bias.
\noindent
We configure the temperature parameter of the InfoNCE loss \cite{oord2018representation} at 0.1, making it learnable to optimize performance. The model is trained over 50 epochs with an early stopping criterion set at a patience of 10. We use the AdamW optimizer \cite{zhuang2022understanding}, with a weight decay of 0.1 for pretraining and 0.02 for fine-tuning. We fine-tune only the projector layer and classifier head for linear evaluation and fine-tune the 'qkv' weight of the last few transformer self attention layers for downstream tasks.
\noindent
Training leverages Automatic Mixed Precision (AMP) in PyTorch for enhanced speed and efficiency with minimal accuracy loss \cite{micikevicius2017mixed}. Instead of using MoCo \cite{he2020momentum} for larger batch sizes, we implement gradient accumulation over 4 steps, which has been shown to be effective in contrastive learning, simulating larger batch sizes \cite{gao2023scaling}. The initial batch size is set at 100, the model is trained on a single A100-SXM4-80GB GPU.\\
\vspace{-8.0mm}
\section{Zero-Shot Processs}
\label{appendix:zero-shot}
The Zero-Shot Classfication process is described below in the figure.
\begin{figure}[!ht]
\centering
\includegraphics[width=\linewidth]{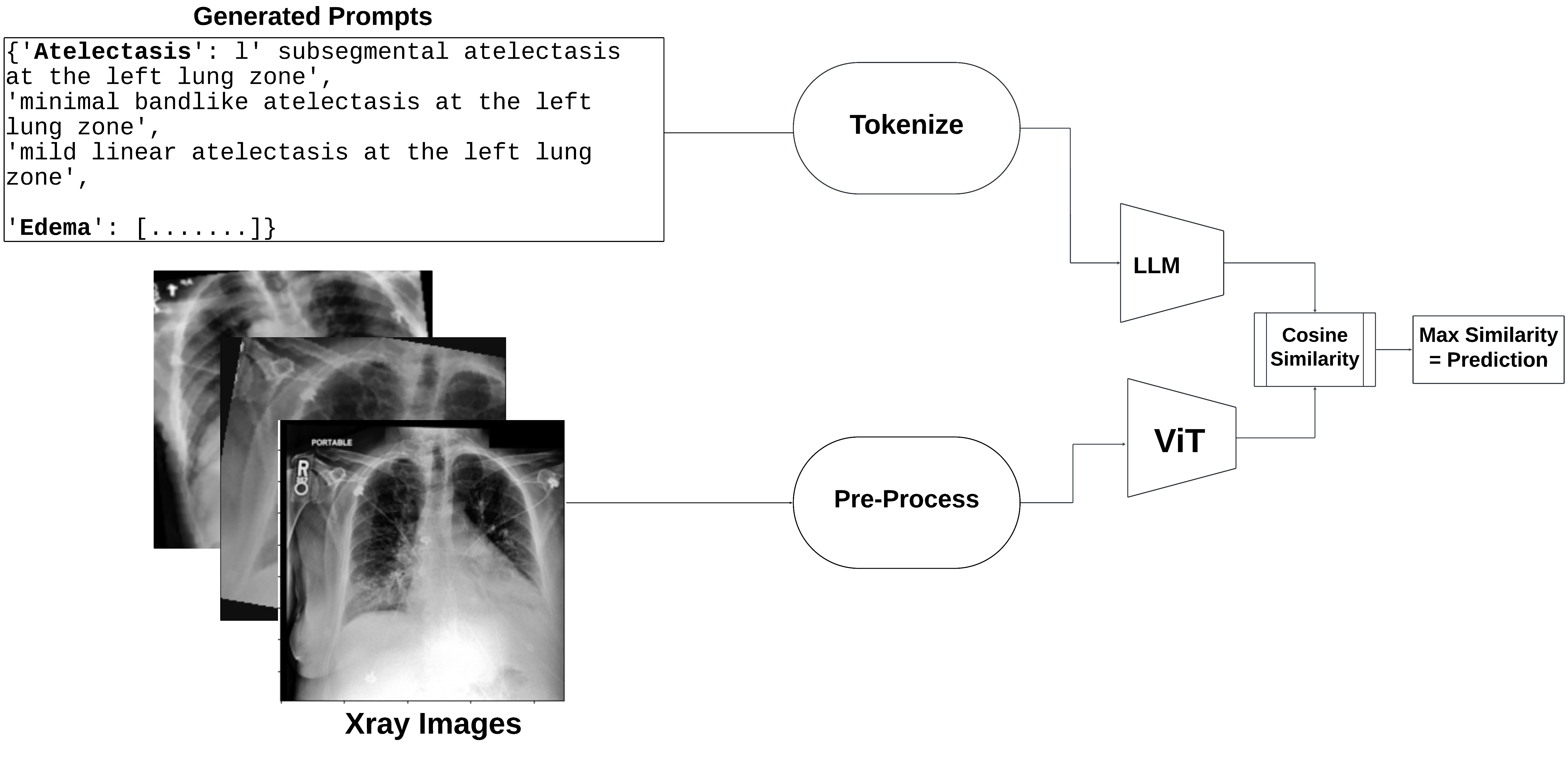}
\caption{Zero Shot Classification Process}
\label{fig:zero-shot}
\end{figure}

\clearpage

\end{document}